\documentclass[12pt]{article}

\usepackage[T2A]{fontenc}
\usepackage[utf8]{inputenc}
 
\usepackage{hyphenat}
\usepackage{xcolor}
\usepackage{hyperref}
\definecolor{linkcolor}{HTML}{000000}
\definecolor{citecolor}{HTML}{0B6B1A}
\definecolor{urlcolor}{HTML}{0B2A5B}
\hypersetup{pdfstartview=FitH, citecolor=citecolor, linkcolor=linkcolor, urlcolor=urlcolor, colorlinks=true}
\usepackage[pdftex]{graphicx}
\graphicspath{{.\\}}
\DeclareGraphicsExtensions{.pdf,.png,.jpg}

\usepackage{authblk}

\usepackage{array}
\usepackage{appendix}

\begin{document}

\title{\bf Stabilizing Elastic Weight Consolidation method in practical ML tasks and using weight importances for neural network pruning}
\author[ ]{Alexey Kutalev$^1$ and Alisa Lapina$^2$}
\affil[1]{SberDevices, PJSC Sberbank, Moscow, Russia}
\affil[2]{NeuroLab, PJSC Sberbank, Moscow, Russia}
\affil[ ]{\it kutalev@gmail.com, ahm.alisa@gmail.com}

\date{\today}


\maketitle

\begin{abstract}
This paper is devoted to the features of the practical application of the Elastic Weight Consolidation (EWC) method for continual learning of neural networks on several training sets. We will more rigorously compare the well-known methodologies for calculating the importance of weights used in the EWC method. These are the Memory Aware Synapses (MAS), Synaptic Intelligence (SI) methodologies and the calculation of the importance of weights based on the Fisher information matrix from the original paper on EWC. We will consider these methodologies as applied to deep neural networks with fully connected and convolutional layers, find the optimal hyperparameters for each of the methodologies, and compare the results of continual neural network  learning using these hyperparameters. Next, we will point out the problems that arise when applying the EWC method to deep neural networks with convolutional layers and self-attention layers, such as the "gradient explosion" and the loss of meaningful information in the gradient when using the constraint of its norm (gradient clipping). Then, we will propose a stabilization approach for the EWC method that helps to solve these problems, evaluate it in comparison with the original methodology and show that the proposed stabilization approach performs on the task of maintaining skills during continual learning no worse than the original EWC, but does not have its disadvantages. In conclusion, we present an interesting fact about the use of various types of weight importance in the problem of neural network pruning.
\end{abstract}

\section{Introduction}

Many scientific works devoted to the topic of the catastrophic forgetting in neural networks, and the detailed description of the problem and attempts to overcome it can be studied from them. The most striking examples of such works from a fairly distant past to the present day can be found in the following articles \cite{c1}-\cite{c4}. 

In 2017, the Elastic Weight Consolidation (EWC) method  \cite{c5} was discovered, which showed outstanding results in overcoming catastrophic forgetting on several classical machine learning tasks in sequential training. The essence of the method is to calculate the importance of each weight (parameter) of the neural network with respect to the tasks that the neural network has already been trained, and to remember the values of the weights after training these tasks. When training subsequent tasks, the method does not allow each weight (parameter) of the neural network to move significantly away from the memorized value with a force that is proportional to the importance of the weight.

Subsequently, several papers have appeared that further refine \cite{c6} and develop \cite{c7}, \cite{c8}, \cite{c9}, \cite{c10} the methodology proposed in \cite{c5}. 
There are also examples of applications of the Elastic Weight Consolidation method in applied tasks \cite{c11}, \cite{c12}, \cite{c13} and its comparative evaluations for different neural network architectures \cite{c12}.

In this paper, we will try to investigate the application of Elastic Weight Consolidation method to some practical machine learning tasks in more detail, discuss the problems encountered and consider the ways to solve them.

\section{Selection of the optimal EWC hyperparameter for specific tasks}

The retention of previously learned knowledge in sequential learning is implemented in the method of Elastic Weight Consolidation by adding a regularizer to the loss function that prevents the most important weights from deviating far from the consolidated values during the learning of subsequent tasks:
\begin{equation}
\label{eq1}
L = L_A + \frac{\lambda}{2} \sum_i \Omega_i (w_i - w^*_i)^2,
\end{equation}
where $L_A$ is the loss function of training {\bf A}, $w^*_{i}$ -- $i$-th weight (parameter) of neural network after training to previous tasks, $\Omega_i$ -- importance of $i$-th weight of neural network after training to previous tasks. From the formula we can see that the contribution of regularizing component to the anti-gradient will be $-\lambda \Omega_i (w_i - w^*_i)$, and thus, when using gradient methods for training, the resistance to change each weight will be proportional to its importance and hyperparameter $\lambda$.

Then if $\lambda$ is small, the resistance to change weights in general will be small, and during sequential learning the neural network will learn the current task better, but it will forget the knowledge from previous learned tasks faster. Conversely, if $\lambda$ is too large, then the resistance to change weights will also be large, and the network will retain the previous knowledge learned on past tasks well, but the learning rate on the current task may be insufficient. The average \textit{accuracy} on all learned tasks can be considered as a quality metric for the sequential learning process.

From this reasoning, it follows that for a particular neural network architecture and a particular set of datasets in sequential learning, there exists an optimal value of $\lambda$, at which the maximum average accuracy after sequential training of all datasets is achieved. And this optimal value $\lambda$ can be found empirically. For example, by a simple grid search.

Accordingly, for the practical application of the Elastic Weight Consolidation method, the task of selecting the optimal hyperparameter $\lambda$ becomes very important to maximize the benefits of the method. In the main papers on the EWC method \cite{c5}, \cite{c7}, \cite{c8}, \cite{c9}, the $\lambda$ selection methodology is either mentioned in passing or not mentioned at all. The accompanying code for these articles also does not contain a full-fledged $\lambda$ selection mechanism. In this paper, we want to correct this omission and try to establish the process of $\lambda$ selection more accurately and strictly.

Further, we decided to compare all methods for computing the weight importances under the same conditions with the optimal value of the hyperparameter $\lambda$ selected for each method. Our experiments were performed for a deep neural network with multiple fully connected layers and for a deep convolutional network. The selection of the optimal $\lambda$ was performed by brute-force over the grid with 20 experiments at each point, calculating the average accuracy and its confidence interval. The capacity (number of parameters) of the neural network was chosen so that all tasks during sequential learning did not fit into the network and there was at least partial crowding out of knowledge of previous tasks on subsequent tasks. The comparing methods of importance calculation were MAS from \cite{c8}, SI from \cite{c7} and method based on Fisher information matrix from \cite{c5}. A detailed description of the experiments can be found in Appendix A.

\begin{figure}[ht]
\centering
\includegraphics[width=1\textwidth]{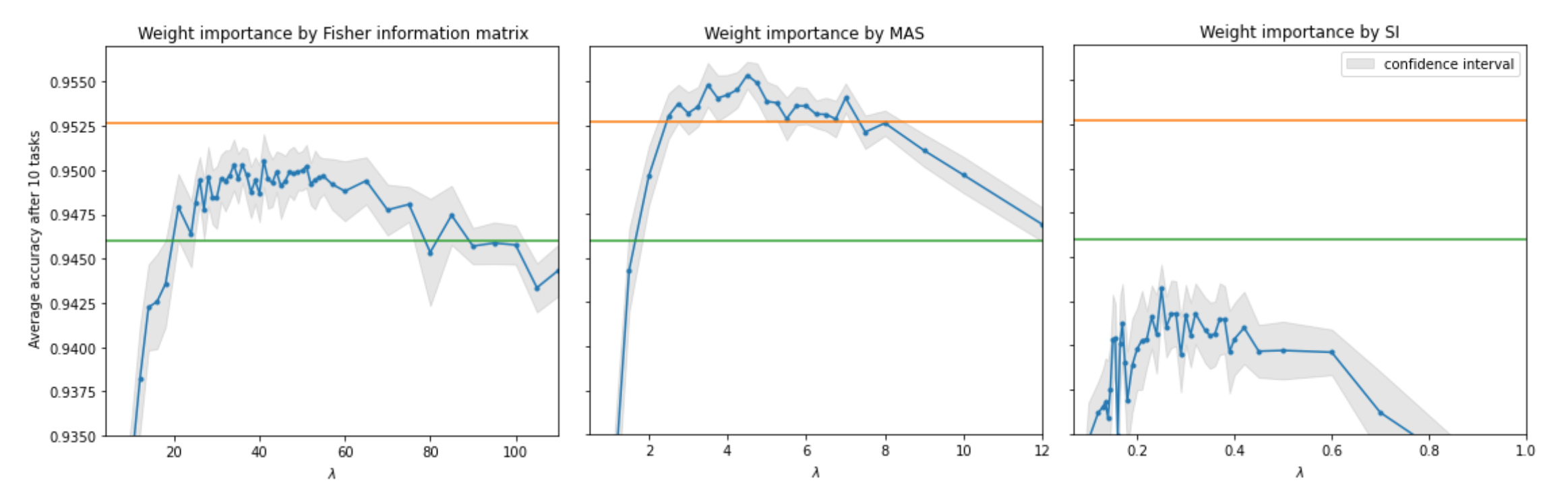}
\caption{Graphs of the dependence of achievable average accuracy for all learned tasks on the hyperparameter $\lambda$ when using importance calculation methods based on the Fisher information matrix, MAS and SI for a network with fully connected layers.}
\label{figure:1}
\end{figure}

\begin{figure}[ht]
\centering
\includegraphics[width=1\textwidth]{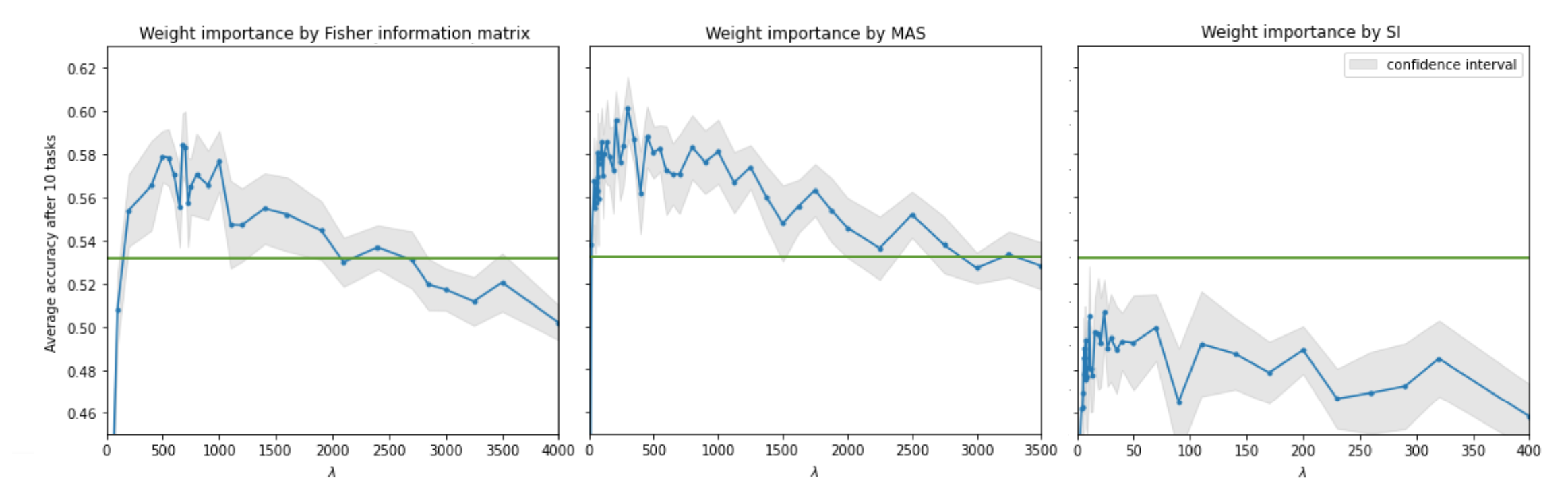}

\caption{Graphs of the dependence of the achievable average accuracy for all learned tasks on the hyperparameter $\lambda$ when using importance calculation methods based on the Fisher information matrix, MAS and SI for a network with convolutional layers.}
\label{figure:2}
\end{figure}

The results of our experiments are shown in Table \ref{table:1} and Figures \ref{figure:1} and \ref{figure:2}.

\begin{table}[ht]
\centering
\begin{tabular}{ m{9em} m{6em} m{12em} }
\hline
Weight importance calculation method & Optimal value of $\lambda$ & Average accuracy and confident interval for this $\lambda$ \\ \hline
\multicolumn{3}{c}{ ~ } \\ 
\multicolumn{3}{c}{\textit{The network with fully connected layers:}} \\
\hline
Fisher & 41 & 0.9505 $\pm$ 0.0015 \\ 
MAS & 4.5 & 0.9553 $\pm$ 0.0008 \\ 
SI & 0.25 & 0.9432 $\pm$ 0.0014 \\
\hline
\multicolumn{3}{c}{ ~ } \\ 
\multicolumn{3}{c}{\textit{The network with convolutional layers:}} \\
\hline
Fisher & 675 & 0.5846 $\pm$ 0.0144 \\ 
MAS & 300 & 0.6012 $\pm$ 0.0144 \\ 
SI & 24  & 0.5068 $\pm$ 0.0150 \\
\hline
\end{tabular}
\caption{Found optimal $\lambda$ values and confidence intervals of the achieved average accuracy at these $\lambda$.}
\label{table:1}
\end{table}

As we see, for a fully connected neural network, the confidence intervals of the average accuracy on the optimal values of $\lambda$ for each of the methods are strictly separated and do not overlap. Thus, we can argue that for the chosen neural network architecture with fully connected layers and sequence of tasks for training, the method of calculating the importance of Memory Aware Synapses (MAS) from \cite{c8} outperforms the method of calculating the importance based on diagonal elements of the Fisher information matrix from \cite{c5}, which in turn outperforms the Synaptic Intelligence (SI) from \cite{c7} in the resulting estimate (average accuracy).

For a neural network with convolutional layers, we see that although the average accuracy values at the optimal values of $\lambda$ form the same order (the average accuracy with MAS weight importances is larger than the average accuracy with weight importances by Fisher information matrix, which, in turn, is larger than the average accuracy with importances by SI), the confidence intervals for average accuracy values for importances by MAS and by Fisher information matrix overlap. Therefore, we cannot claim in a strict sense superiority of the MAS importance calculation method over the Fisher information matrix importance calculation method. However, we can claim superiority of these methods over the SI method because the confidence intervals for the average accuracy using these methods are strictly separated from the confidence interval for the average accuracy value for the importances calculated by the SI method.

It should be noted that there is a fundamental difference between the \cite{c5} (based on diagonal elements of the Fisher information matrix) and the \cite{c7}, \cite{c8}, \cite{c9} methods of weight importance calculation. The method from \cite{c5} (the original EWC) uses the output values of the task training dataset , and thus tries to retain that part of the generalizations/representations embedded in the training dataset of the task that the neural network was able to learn. The methods from \cite{c7}, \cite{c8}, \cite{c9} do not use the output values of the training dataset, and thus try to preserve the generalizations/representations that the neural network has actually learned.

The method from \cite{c5} can also be used to store the actual learned knowledge. For this purpose, the output of the neural network with the maximum value should be taken when calculating the importance on each example of dataset, or the output should be selected by random sampling from the output distribution. This approach is used in most of the EWC implementation examples available on the Internet. However, when applying the method from \cite{c5} to large NLP models (e.g., BERT, GPT, etc.), it looks more preferable to use the method with output selection for importance calculation according to the output values from the training dataset, since in such cases the goal is usually to preserve with EWC the natural language knowledge contained in the texts presented during training, rather than the language representations that the model was able to learn.

\section{Problems of the Elastic Weight Consolidation method}

Applying the method of Elastic Weight Consolidation to networks with convolutional layers and networks with self-attention layers, we found that, in contrast to networks with fully connected layers, the calculated importance of weights in convolutional layer or layer with self-attention have a very large variance. That is, there appear to be weights (parameters) of the network whose importance is several orders of magnitude greater than the average importance of the layer. The effect of such importance can have a significant impact on the effectiveness of the method. To clarify this effect let us consider the process in more detail.

As we mentioned earlier, the regularizing component of the loss function when applying EWC gives a contribution to the anti-gradient of the form $-\lambda \Omega_i (w_i - w^*_i)$. When using as an optimizer, for example, the stochastic gradient descent method, the weight increment at the learning step will be of the form:
$$
\Delta w_i = -\alpha\frac{\partial L_A}{\partial w_i} - \alpha\lambda \Omega_i (w_i - w^*_{A,i}),
$$
where $\alpha$ is the learning rate, $L_A$ is the loss function for the current task {\bf A}, $\lambda$ -- hyperparameter of the EWC method, $\Omega_i$ -- importance of $i$-th weight of the neural network, $w_i$ is the current value of $i$-th weight, $w^*_i$ -- value of $i$-th weight after the previous task training is completed.
In case the importance of the $i$-th weight is large enough, the contribution from the loss function $-\alpha\frac{\partial L_A}{\partial w_i}$ to the weight increment will be negligible. Further, if $\Omega_i \geq \frac{1}{\alpha\lambda}$, then after the optimization step the weight $w_i$ will not shift towards $w^*_i$ but will jump over it. 
In the case $\Omega_i \geq \frac{2}{\alpha\lambda}$, the weight $w_i$ will not only jump over the value of $w^*_i$, but the distance between $w_i$ and $w^*_i$ will increase by a factor of $(\alpha\lambda\Omega_i - 1)$. Thus, there will be an effect known as ``gradient explosion''. And if even with this ``explosion'' the learning process will not be interrupted by overflow, the most important weights for the previous tasks will be moving very fast away from the consolidated values.

Of course, the use of optimization methods with momentum or other compensating mechanisms can prevent the described effect, but even in this case it can negatively affect the convergence of the optimization method and the ability of the EWC method to retain previously learned knowledge.

There is also negative effect of the extra-large importances of weights while using gradient clipping. These weights with extra-large importances may give so big contribution to gradient norm that after gradient clipping the contribution of the other weights become vanishingly small. Consequently, this clipped gradient serves mainly to bring these hyper-important weights back to their consolidated values, and information from the current batch of samples that is useful for learning is largely ignored.

To solve these problems, we used a stabilization mechanism that prevents the appearance in the regularizing contribution to the weight increment of values larger than the difference of the consolidated and current weight values. For this purpose we transformed the loss function to the form:
$$
L = L_A + \frac{\lambda}{2} \sum_i \frac{\Omega_i}{\alpha\lambda\Omega_i + 1} (w_i - w^*_i)^2.
$$
For such loss function the contribution of the regularizing component to the anti-gradient will be
$$
-\frac{\alpha\lambda\Omega_i}{\alpha\lambda\Omega_i + 1} (w_i - w^*_{A,i}).
$$
At small (close to zero) importance of $\Omega_i$ the multiplier $\frac{\alpha\lambda\Omega_i}{\alpha\lambda\Omega_i + 1}$ behaves proportional to $\alpha\lambda\Omega_i$, and at very large $\Omega_i$ tends to 1.
Thus the contribution to the anti-gradient of the regularizing component will not exceed the difference of weights even at any large importance of weight.

\begin{table}[ht]
\centering
\begin{tabular}{ m{8em} m{5em} m{7.5em} m{7.5em} }
\hline
Method & Importance & Optimal $\lambda$ & Confidence Interval \\ \hline
\multicolumn{4}{c}{ ~ } \\ 
\multicolumn{4}{c}{\textit{Neural network with fully connected layers:}} \\ \hline
EWC & Fisher & 41 & 0.9505 $\pm$ 0.0015 \\ 
stabilized EWC & Fisher & 85 & 0.9510 $\pm$ 0.0011 \\ \hline
EWC & MAS & 4.5 & 0.9553 $\pm$ 0.0008 \\ 
stabilized EWC & MAS & 8.5 & 0.9554 $\pm$ 0.0009 \\ \hline
EWC & SI & 0.25 & 0.9432 $\pm$ 0.0014 \\
stabilized EWC & SI & 0.64 & 0.9422 $\pm$ 0.0017 \\ \hline
\multicolumn{4}{c}{ ~ } \\ 
\multicolumn{4}{c}{\textit{Neural network with convolutional layers:}} \\
\hline
EWC & Fisher & 675 & 0.5846 $\pm$ 0.0144 \\ 
stabilized EWC & Fisher & 1300 & 0.5872 $\pm$ 0.0161 \\ \hline
EWC & MAS & 300 & 0.6012 $\pm$ 0.0144 \\ 
stabilized EWC & MAS & 450 & 0.5930 $\pm$ 0.0149 \\ \hline
EWC & SI & 24 & 0.5068 $\pm$ 0.0150 \\
stabilized EWC & SI & 140 & 0.5106 $\pm$ 0.0188 \\ \hline
\end{tabular}
\caption{Comparing the optimal $\lambda$ values and confidence intervals of the resulting average accuracy when using the original and stabilized method of Elastic Weight Consolidation.}
\label{table:2}
\end{table}

\begin{figure}[p]
\centering
\includegraphics[width=1\textwidth]{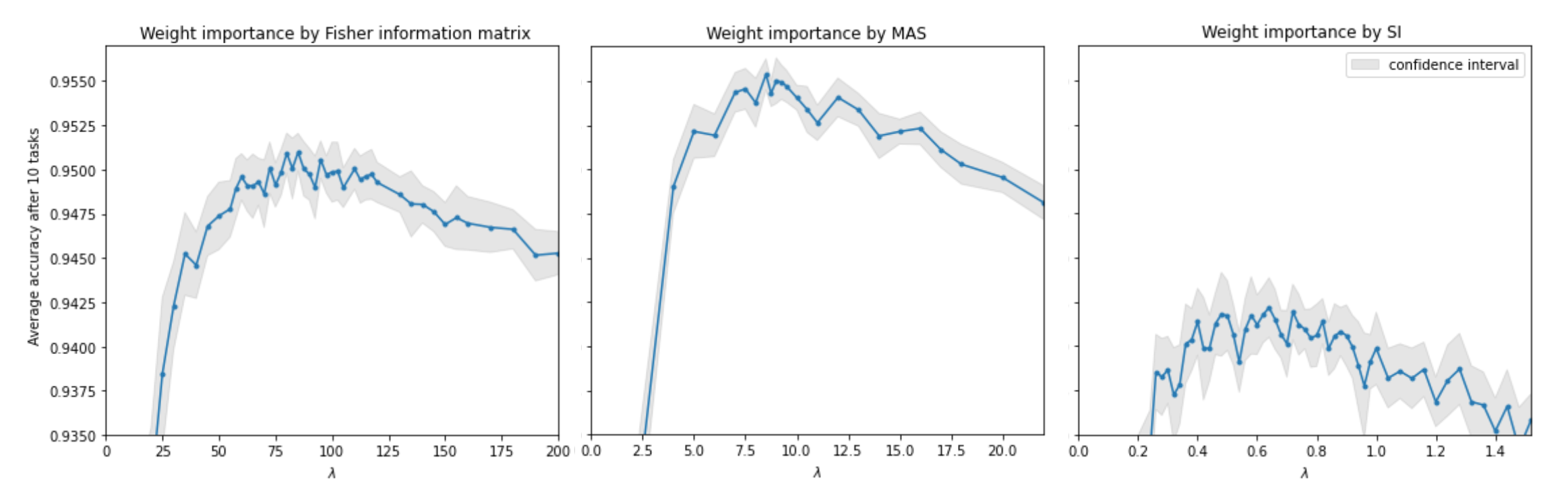}
\caption{Graphs of the dependence of achievable average accuracy for all learned tasks on the hyperparameter $\lambda$ when using importance calculation methods based on the Fisher information matrix, MAS and SI for a network with fully connected layers.}
\label{figure:3}
\end{figure}

\begin{figure}[p]
\centering
\includegraphics[width=1\textwidth]{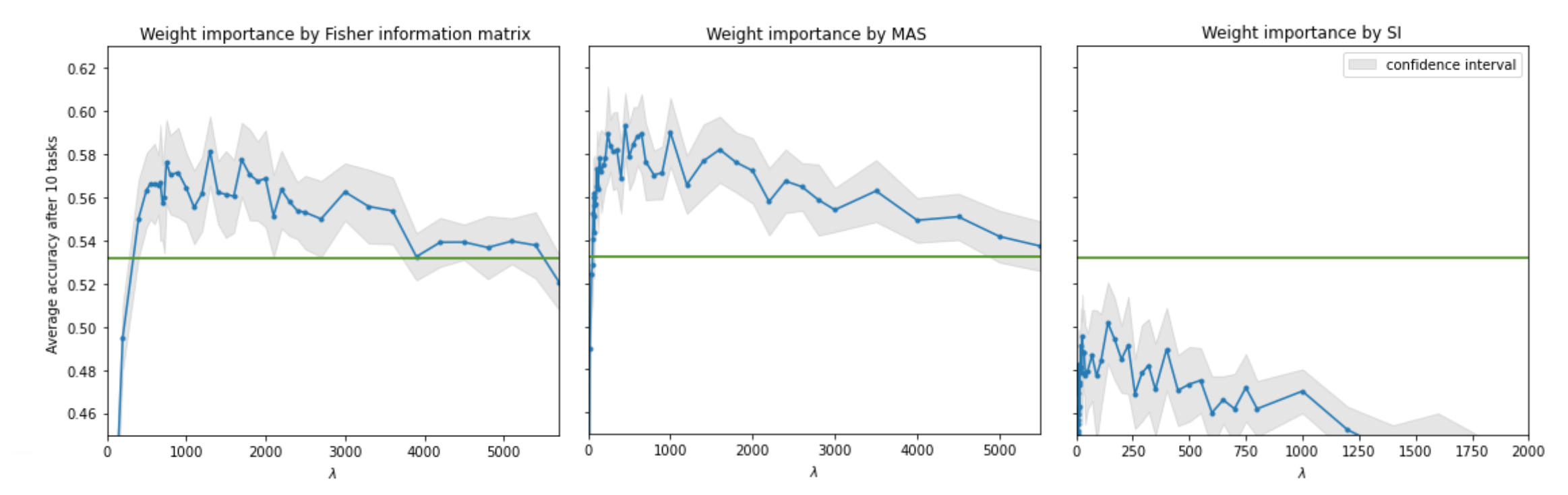}
\caption{Graphs of the dependence of the achievable average accuracy for all learned tasks on the hyperparameter $\lambda$ when using importance calculation methods based on the Fisher information matrix, MAS and SI for a network with convolutional layers.}
\label{figure:4}
\end{figure}

To test the workability of this approach, we performed experiments similar to those in Section 2 to verify that it, like the original Elastic Weight Consolidation method, allows successful knowledge retention during sequential learning, and also allows for no worse average accuracy than EWC at the optimal $\lambda$.

Table \ref{table:2} and Figures \ref{figure:3} and \ref{figure:4} show the experimental results and their comparison at optimal $\lambda$ with the original EWC results.

As we see, both for the networks with fully connected layers and for convolutional networks when using the Elastic Weight Consolidation method with the proposed stabilization mechanism, the average accuracy values at the optimal hyperparameters $\lambda$ are almost always higher than the average accuracy values when training using the Elastic Weight Consolidation method without stabilization. However, the confidence intervals for the average accuracy values for the cases with and without stabilization always overlap, i.e., we cannot speak with certainty about the superiority of one of the approaches. Thus, the proposed stabilization mechanism at least does not worsen the method of Elastic Weight Consolidation, while improving the convergence of optimization algorithms and preserving previously learned knowledge.

We also conducted experiments on the application of a stabilized version of the EWC method to the fine-tuning of the pretrained Russian GPT2 model on the dialog training dataset. As a result, the gain from using the method of EWC was about 7 units of perplexity compared to conventional fine-tuning (the model's perplexity was calculated on the set of texts from books, it was about 16 on the pretrained model, after conventional fine-tuning it became 26, and when fine-tuning with stabilized EWC it became 19).

\section{Neural network pruning using weight importances of Elastic Weight Consolidation method}

Since the importance of weights in the Elastic Weight Consolidation method is positioned as a measure of how important each individual weight is for retaining learned representations, it is logical to try to use the values of these importance for the task of network pruning in order to reduce the size and computational complexity of the neural network (\cite{c16}-\cite{c21}).

\begin{figure}[ht]
\centering
\includegraphics[width=1\textwidth]{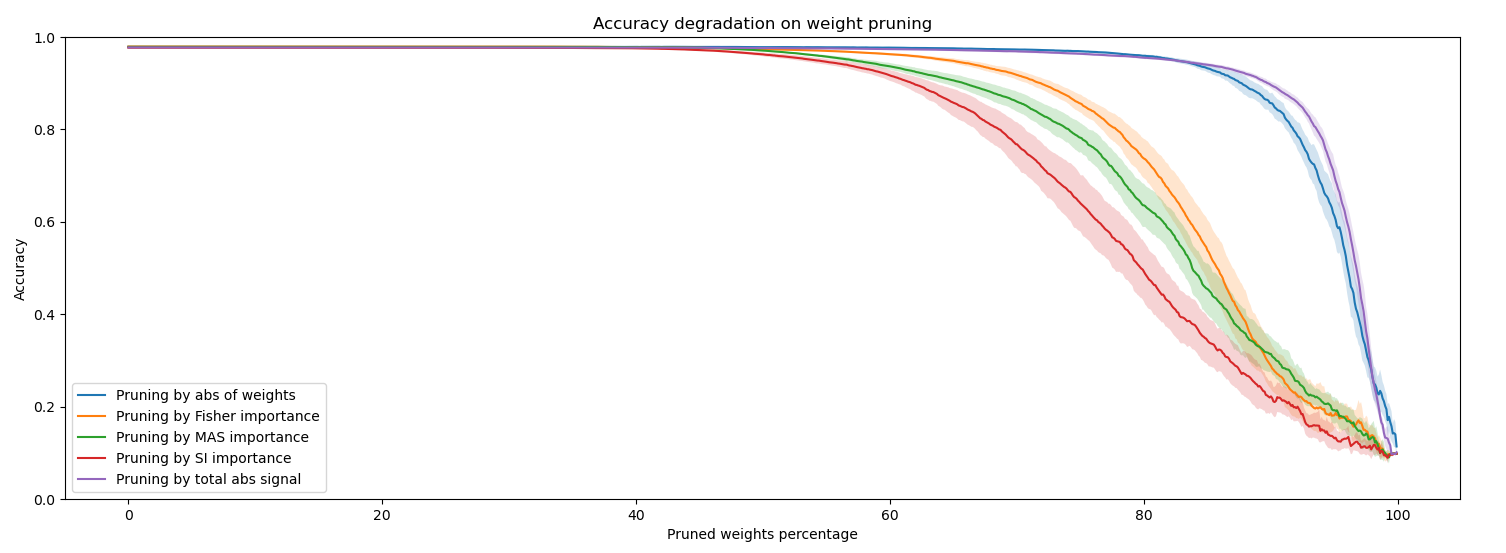}

\caption{Graphs of average accuracy degradation as a function of the number of pruned (zeroed) weights of the neural network. The average accuracy over 10 runs is given along with its confidence interval.}
\label{figure:5}
\end{figure}

Accordingly, we performed unstructured pruning experiments in a neural network based on EWC weight importances computed in different ways, and compared the results with pruning based on the weight magnitude. The results are shown in Figure \ref{figure:5}.

As can be seen in the graph, importances by diagonal elements of Fisher information matrix \cite{c5}, and the SI \cite{c7} and MAS \cite{c8} importances performed much worse in the pruning task than weight importances based on the total absolute signal \cite{c9} and weight magnitude.

This is very unexpected result. As the explanation we can suppose that weight importance value by SI, MAS and Fisher describes importance in a small neighborhood of the value of weight. But pruning by zeroing weight moves the weight value outside this neighborhood, so weight importance value loses its meaning.

\section{Conclusions}

The presented methodology for finding the hyperparameter $\lambda$ for the Elastic Weight Consolidation method allows us to find $\lambda$ that is close enough to the optimal one for a particular neural network and machine learning task.

As a result of its application, we found that for all neural network architectures used in our experiments, the choice of MAS method for calculating the weight importances is optimal, as it contributes to the best knowledge retention in terms of average accuracy in sequential learning. That is, MAS is statistically significantly better than SI for all used architectures, and statistically significantly better than Fisher information matrix calculation of weight importances for the network with fully connected layers.

The proposed approach of stabilizing the method of Elastic Weight Consolidation preserves its ability to overcome catastrophic forgetting and shows results no worse (in terms of average accuracy) than the original method.

In the task of unstructured pruning (pruning of connections by zeroing the weights) in a neural network of fully connected layers the optimal is pruning by weight magnitude or by the total absolute signal passed through the connection. Both of these methods are significantly superior to pruning based on MAS, SI and Fisher Information Matrix importances.

Since most of modern language models use multilayer neural networks based on self-attention (transformers), we find very promising the use of stabilized method of Elastic Weight Consolidation in fine-tuning of pretrained models on specialized language datasets.


\begin{appendices}

\section{Appendix}

\begin{figure}[ht]
\centering
\includegraphics[width=1\textwidth]{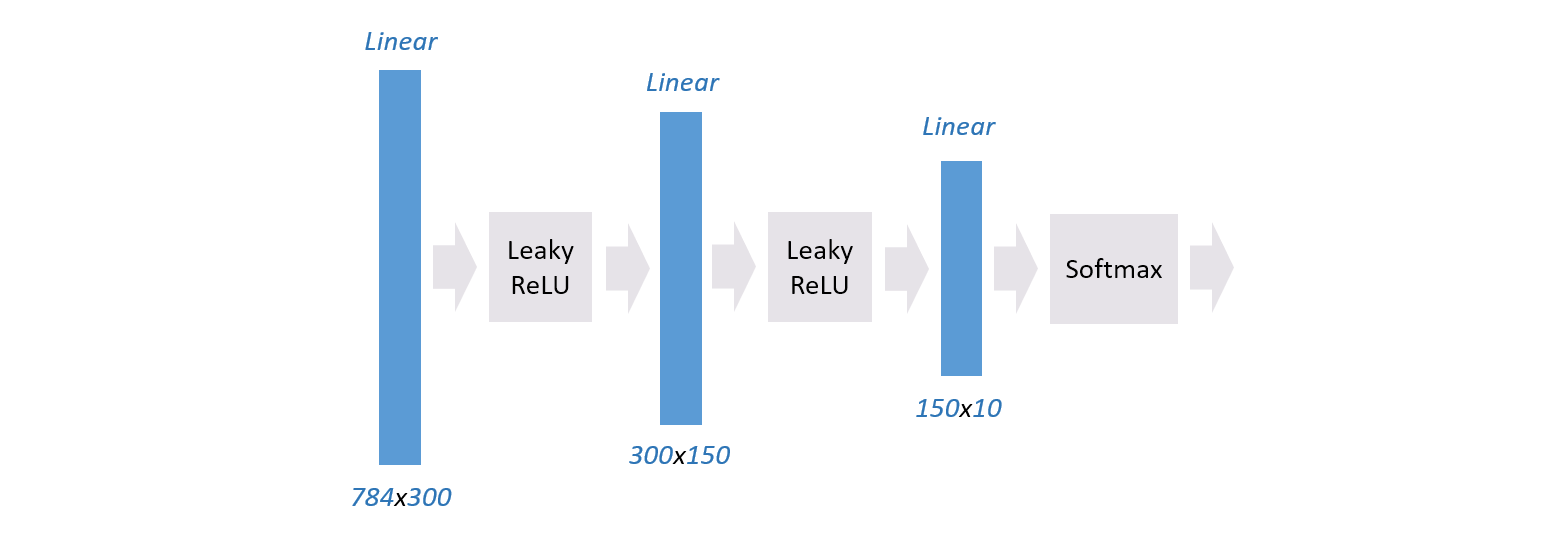}
\caption{Structure of neural network with fully connected layers}
\label{figure:6}
\end{figure}

\begin{figure}[ht]
\centering
\includegraphics[width=1\textwidth]{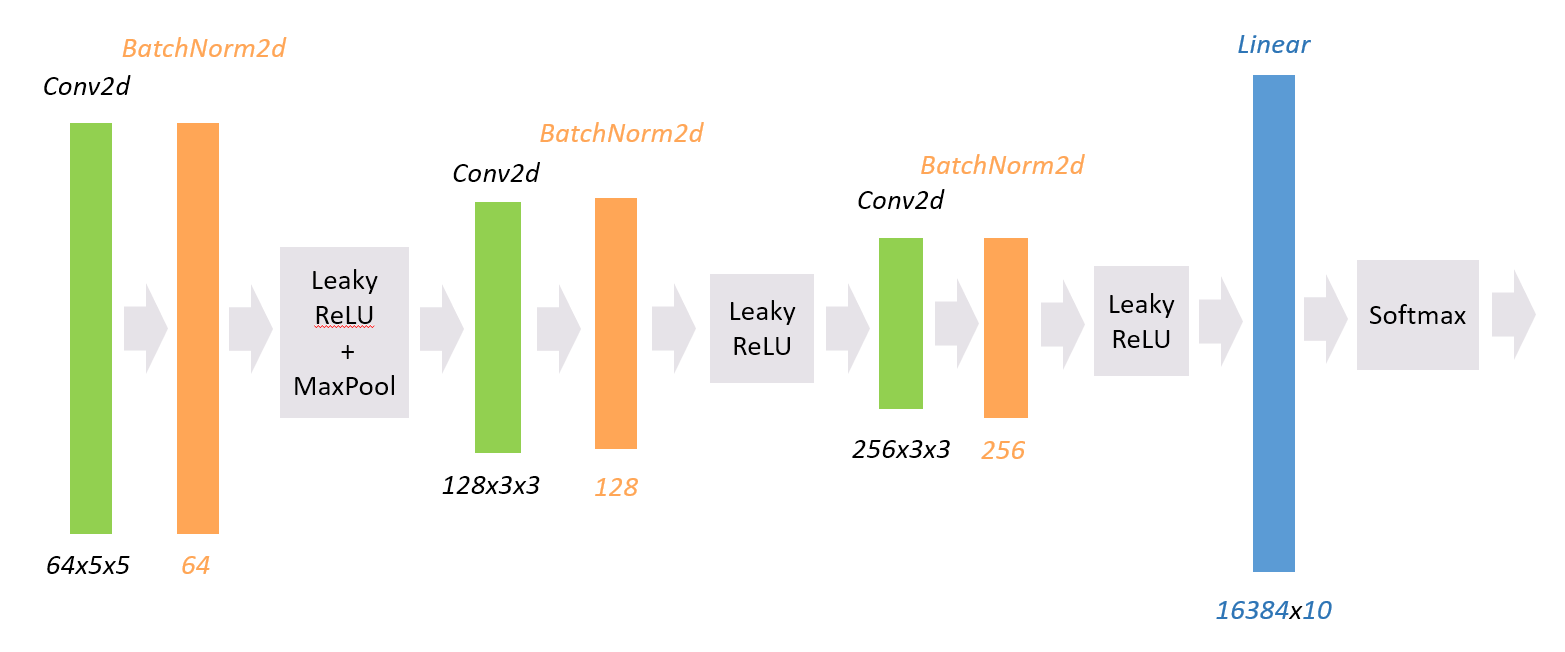}
\caption{Structure of neural network with convolutional layers}
\label{figure:7}
\end{figure}

The architecture of the neural network with fully connected layers used in our experiments is shown in Figure \ref{figure:6}, the architecture of the network with convolutional layers is shown in Figure \ref{figure:7}.

Ten training datasets obtained from MNIST dataset by random permutations of values in the input vector of each example were used for sequential training of the  network with fully connected layers (first a random permutation in the input vector was generated, then the same permutation was applied to all examples from MNIST, and thus ten training datasets were generated by ten random permutations of inputs).

The training datasets MNIST and FasionMNIST and their $\pi/2$ rotations were used for sequential training of the convolutional network. In total, 4 training datasets participated in the sequential training of the convolutional network.

In a single experiment, the neural network was trained by Adam method with parameters $learning~rate = 0.001$, $\beta_1=0.9$, $\beta_2=0.999$, $\epsilon=10^{-8}$. Prior to experiment start, the values of all weight importances in the network were initialized to zeros. Training was performed sequentially on all training datasets (10 for the network with fully connected layers, 4 for the network with convolutional layers). Training for each training dataset was performed for 6 epochs with minibatch size of 100. After training on the training dataset, the weight importances on this dataset were calculated and these weight importances were summed up with the weight importances obtained earlier. 
After training all datasets in the experiment, the \textit{accuracy} was measured for all delayed (test) parts of the training datasets learned by the network. The obtained accuracy was used as the result of the experiment.

For each value of the hyperparameters, the experiment was conducted twenty times. The result was averaged and the confidence interval for the mean was calculated with statistical significance of $0.95$.

The code used for the experiments is available  \texttt{\href{https://github.com/aakutalev/ewc-features}{here}}.

\end{appendices}

\end{document}